\pdfoutput=1
\documentclass[10pt,twocolumn,letterpaper]{article}

\usepackage[pagenumbers]{iccv} 
\usepackage{lipsum}
\newcommand\blfootnote[1]{%
  \begingroup
  \renewcommand\thefootnote{}\footnote{#1}%
  \addtocounter{footnote}{-1}%
  \endgroup
}

\definecolor{iccvblue}{rgb}{0.21,0.49,0.74}
\usepackage[pagebackref,breaklinks,colorlinks,allcolors=iccvblue]{hyperref}

\def\paperID{1885} 
\def\confName{ICCV}
\def\confYear{2025}

\title{SwapAnyone: Consistent and Realistic Video Synthesis for Swapping Any Person into Any Video}

\author{Chengshu Zhao$^{1\ast}$ 
\quad
Yunyang Ge$^{1,2\ast}$
\quad
Xinhua Cheng$^{1,2}$
\quad
Bin Zhu$^{1,2}$
\quad
Yatian Pang$^3$\\
Bin Lin$^{1,2}$
\quad
Fan Yang$^1$
\quad
Feng Gao$^{1\dag}$
\quad
Li Yuan$^{1\dag}$ \\
$^1$Peking University \quad $^2$Rabbitpre Intelligence \quad $^3$National University of Singapore\\
}

\begin{document}

\twocolumn[{
\renewcommand\twocolumn[1][]{#1}
\maketitle
\begin{center}
    \captionsetup{type=figure}
    \includegraphics[width=\textwidth]{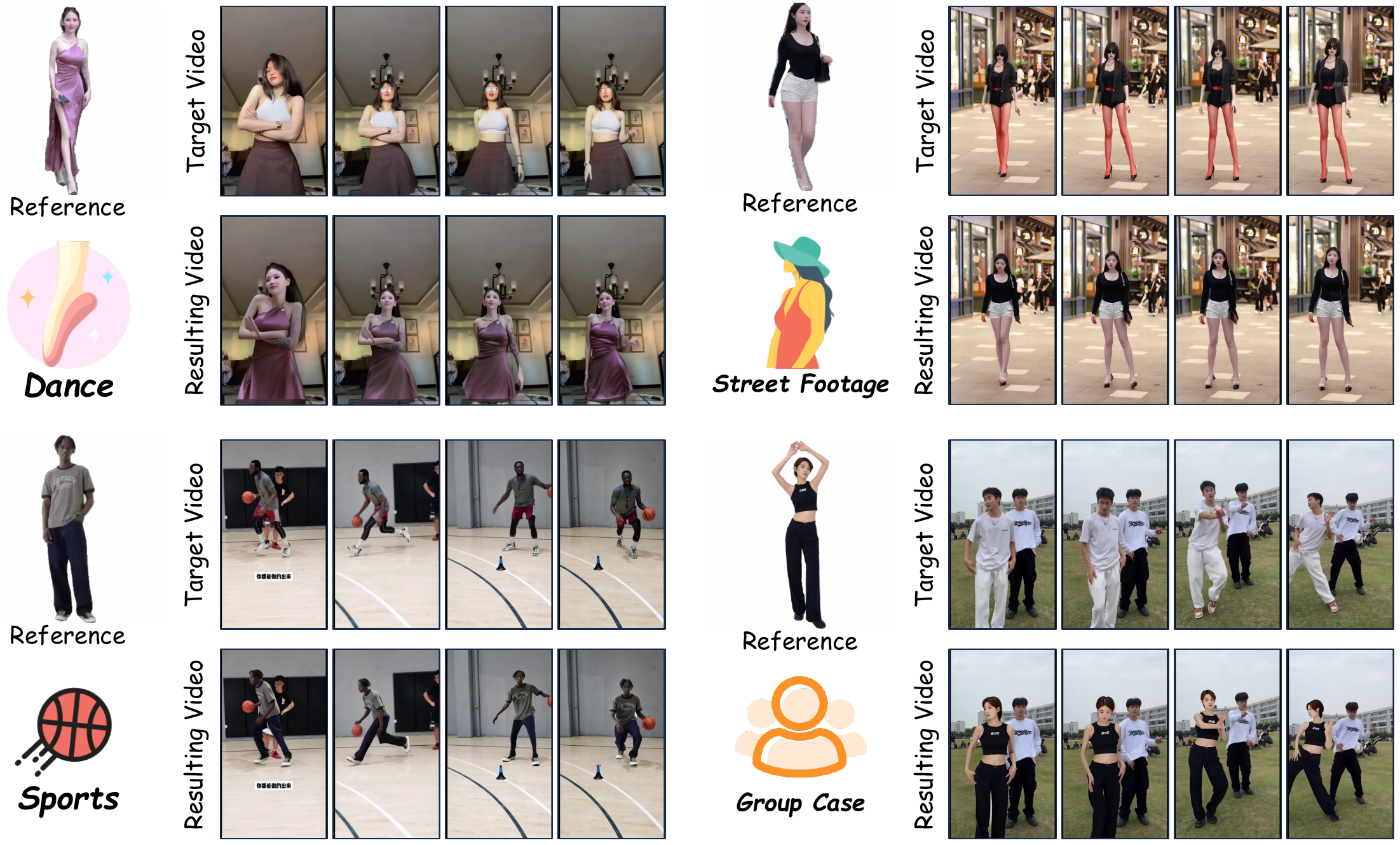}
    \captionof{figure}{SwapAnyone allows users to provide a reference body image and a target video from any source, then seamlessly swap the provided body with the original body in the target video to produce a highly realistic video.}
    \label{fig:teaser}
\end{center}
}]

\blfootnote{$^\ast$ Equal contribution to this work.}
\blfootnote{$^\dag$ Corresponding Author.}

\begin{abstract}
Video body-swapping aims to replace the body in an existing video with a new body from arbitrary sources, which has garnered more attention in recent years. Existing methods treat video body-swapping as a composite of multiple tasks instead of an independent task and typically rely on various models to achieve video body-swapping sequentially. However, these methods fail to achieve end-to-end optimization for the video body-swapping which causes issues such as variations in luminance among frames, disorganized occlusion relationships, and the noticeable separation between bodies and background. In this work, we define video body-swapping as an independent task and propose three critical consistencies: \textbf{identity consistency}, \textbf{motion consistency}, and \textbf{environment consistency}. We introduce an end-to-end model named \textbf{SwapAnyone}, treating video body-swapping as a video inpainting task with reference fidelity and motion control. To improve the ability to maintain environmental harmony, particularly luminance harmony in the resulting video, we introduce a novel EnvHarmony strategy for training our model progressively. Additionally, we provide a dataset named \textbf{HumanAction-32K} covering various videos about human actions. Extensive experiments demonstrate that our method achieves \textbf{State-Of-The-Art (SOTA)} performance among open-source methods while approaching or surpassing closed-source models across multiple dimensions. All code, model weights, and the HumanAction-32K dataset will be open-sourced at \url{https://github.com/PKU-YuanGroup/SwapAnyone}.
\end{abstract}
\section{Introduction}
DeepFake is a fake generation technology using deep neural networks to substitute personal information into a target image or video, as seen in applications like face-swapping~\cite{face2face,deep_video_portraits,faceshifter,simswap,diffface,diffswap}. Recently, with broad application potential in fields such as filmmaking and interactive entertainment, video body-swapping has attracted more attention as an advanced task. In this task, users supply a reference body image and a target video, allowing the model to replace the body in the target video with the reference. However, unlike video face-swapping which focuses on the face region, video body-swapping involves the entire body, demanding the integration of more complex visual information and a deeper understanding of physical interactions between the foreground body and background objects.

With the advancement of video generation models~\cite{animatediff,lavie,stable_video_diffusion,magictime,dreamdance,opensoraplan,opensora,latte,videocrafter,videocrafter2,viewcrafter} and video understanding models~\cite{lin2023video,lin2024moe,zhu2023languagebind,chen2024sharegpt4video,ning2023video,qwenvl}, character-generated technologies have made significant progress. Some works~\cite{dynamicrafter,idanimator,moonshot,consisid,fantasyid} can generate videos with consistent identity based on a reference face or body image, and some~\cite{animateanyone,champ,poseanimate,magicanimate,i2v_adapter,skyreels,i2vcontrol,AniGS} extend this by incorporating motion control. However, these methods still focus on creating a new video rather than altering an existing one. When it comes to modifying a target video, some works~\cite{mimo,moviecharacter} can already replace the body in the target video with the reference. Nevertheless, to the best of our knowledge, current technologies still treat video body-swapping as the combination of two tasks: \textbf{(i)} generating videos conditioned on the identity of reference and the motion of target videos, and \textbf{(ii)} fusing the generated videos with the target videos. As a result, multiple separate models are needed to handle different tasks, which hinders the end-to-end optimization and struggles with issues such as luminance inconsistency between bodies and backgrounds, chaotic occlusion relationships, and background variations.

In this work, we propose treating video body-swapping as an independent task instead of a composite of multiple tasks. We highlight three critical consistencies that lie at the core of the video body-swapping task:
\textbf{(i) Identity consistency.} The identity of the body in the resulting video should match the reference, including facial characteristics, expressions, clothing, body shape, and other recognizable details.
\textbf{(ii) Motion consistency.} The body’s motion in the resulting video should match that of the target. 
\textbf{(iii) Environment consistency.} Environment consistency requires the background of generated video to be identical to that of the target. Moreover, regardless of the luminance of the reference body image from any source, the resulting video must preserve uniform luminance.

For achieving objective of the video body-swapping task and ensuring the required consistencies, we present a novel end-to-end model named \textbf{SwapAnyone}. To ensure environment consistency, we introduce Inpainting UNet based on the Stable-Diffusion-v1.5-Inpainting~\cite{stable_diffusion} model which performs image inpainting by taking the user's input image and mask, then regenerating the masked areas based on text prompts. We replace the text encoder of the model with CLIP~\cite{clip} image encoder to guide inpainting using image semantics. To support video input, Temporal Layers are chosen in SwapAnyone, which are added after spatial attention layers in the Inpainting UNet to enhance video smoothness. To ensure identity consistency, we introduce an ID Extraction Module, which is a complete copy of the UNet used in Stable-Diffusion-v1.5. It takes the reference body image and its corresponding DWpose\cite{dwpose} image as input and injects their features into the Inpainting UNet. To ensure motion consistency, we design Motion Control Module that encodes the DWpose sequence of the target video and injects it into the latents, ensuring the actions in the resulting video closely match those in the target video. Additionally, we propose a luminance-based regularization method named EnvHarmony Strategy, which enables the model to ignore the influence of reference luminance during training and focus on extracting identity features. We also construct a dataset named HumanAction-32K, which covers common scenarios involving human motions.
In summary, the contributions of this work can be summarized as follows:
\begin{itemize}
    \item \textbf{Task Definition.} We define video body-swapping as an independent task and specify its definition along with the three consistencies.
    \item \textbf{An End-to-End Model with EnvHarmony Strategy.} We optimize the video body-swapping task in an end-to-end manner. In this manner, we use the Inpainting UNet and the EnvHarmony strategy to maintain environment consistency, and design other modules to achieve identity consistency and motion consistency.
    \item \textbf{A Task-Specific Dataset.} We present a dataset named HumanAction-32K covering various common human actions, such as dance, sports, street footage, and daily vlogs, which is suitable for video body-swapping task.
\end{itemize}
\section{Related Work}
\subsection{Diffusion for Image and Video Synthesis}
Diffusion-based image synthesis~\cite{ddpm,stable_diffusion,roompainter,chronomagic,wfvae,dit,pixart_alpha,pixart_sigma,prompt2poster} has made significant advancements in recent years. Denoising Diffusion Probabilistic Models (DDPM)~\cite{ddpm} is the first successful example of applying diffusion to image synthesis. Subsequently, the Latent Diffusion Model (LDM)~\cite{stable_diffusion} shifts the diffusion process from pixel space to the latent space of a Variational Autoencoder (VAE)~\cite{vae,wfvae,odvae}. Stable Diffusion is an extension of LDM, which trains on a subset of Laion-5B~\cite{laion}.

In Video diffusion models~\cite{opensora,opensoraplan,video_ldm,animatediff,stable_video_diffusion}, Video LDM~\cite{video_ldm} represents an early attempt by pre-training on images and then fine-tuning with temporal layers on videos. Subsequently, AnimateDiff~\cite{animatediff} introduces a novel approach that allows for converting image diffusion models into video diffusion models by freezing the weights of spatial layers and only fine-tuning added temporal layers. Stable Video Diffusion (SVD)~\cite{stable_video_diffusion} emphasizes the importance of datasets, enabling high-quality synthesis.

\subsection{Controllable Character Video Synthesis}
AnimateAnyone~\cite{animateanyone} is the first to introduce the concept of controllable character video synthesis that utilizes user-provided poses to drive the motion of characters in images~\cite{magicpose,animate_x,dreampose,tcan,ren2020deep,magicanimate}. MimicMotion~\cite{mimicmotion} is based on the SVD~\cite{stable_video_diffusion} and improves the details of the results by introducing confidence-aware pose guidance. It also incorporates a progressive latent fusion strategy to generate longer videos. MIMO~\cite{mimo} decomposes the video into three spatial components (scene, human, and occlusion) and utilizes SMPL~\cite{SMPL} to construct richer control information, enabling the synthesis of highly controllable character videos.
\subsection{Identity Injection and Swapping}
Large-scale pre-trained diffusion models typically support only text conditioning rather than synthesis based on a given reference image, limiting the fidelity of specified identities across different scenarios. IP Adapter~\cite{ipadapter} uses CLIP~\cite{clip} image encoder as a feature extractor, which enables the diffusion model to generate images containing elements from a user-provided reference image. Putting People in Their Place~\cite{putting_people} is trained by extracting two different frames from a video as the reference and target, allowing for the insertion of any human reference into any image. 

In the video domain, several works~\cite{consisid,idanimator,fantasyid} have achieved the synthesis of reference-based videos by leveraging CLIP or specialized feature encoders to extract features from the reference and inject them into the denoiser. In addition to controllable character video synthesis, MIMO~\cite{mimo} can generate body-swapping videos based on a reference image by first performing inpainting on the target videos and then merging the 3D pose-controlled character videos with the inpainted videos. MovieCharacter~\cite{moviecharacter} adopts models across three tasks, which comprise controllable character video synthesis, video inpainting, and video composition to achieve video body-swapping.
\section{Method}
\begin{figure*}[t]
    \centering
    \includegraphics[width=0.95\linewidth]{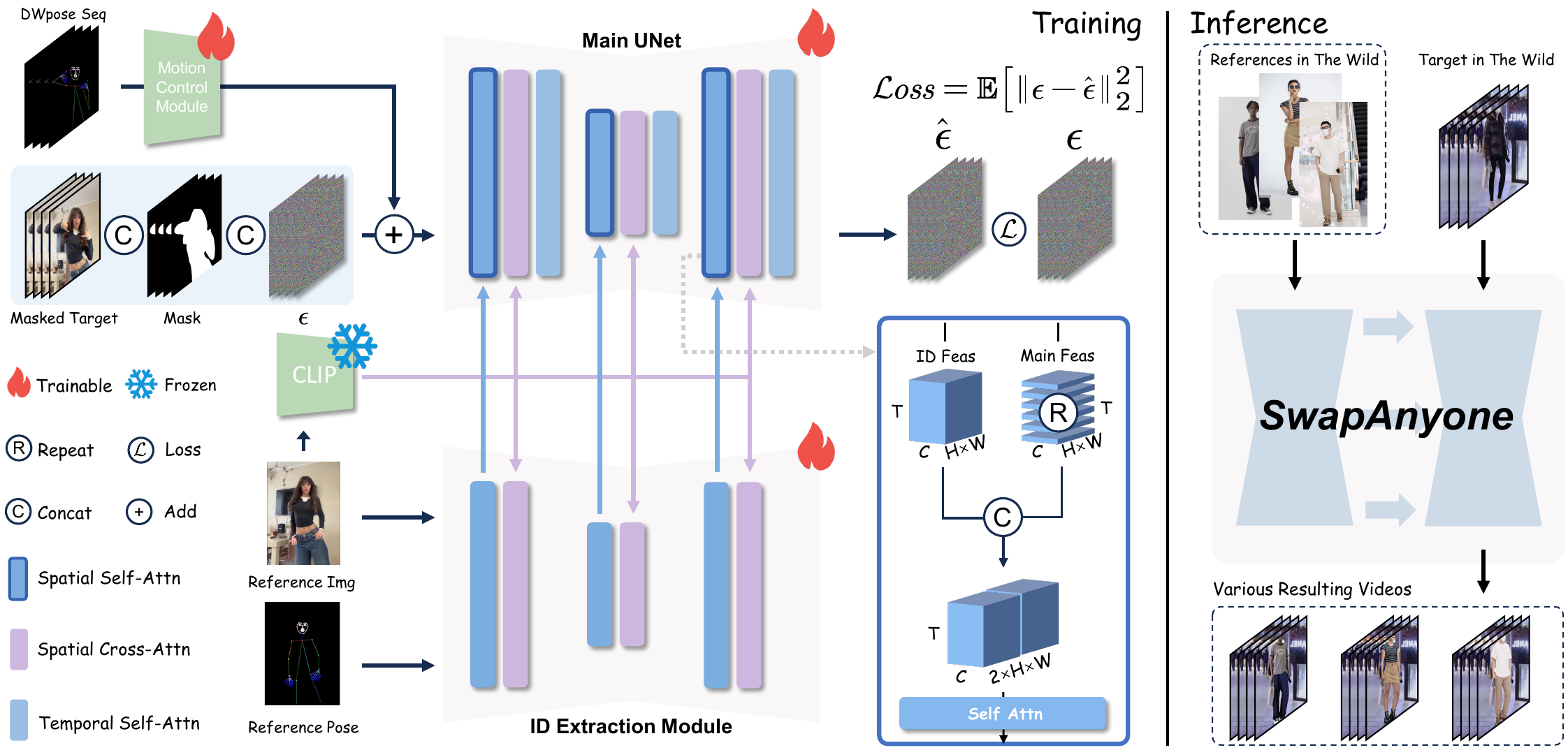}
    \caption{Overview of SwapAnyone. Firstly, the user-provided reference body image and corresponding DWpose image are processed by the ID Extraction Module. Simultaneously, the DWpose sequence of the body in the target video is sent to the Motion Control Module to extract motion features, which are incorporated into the latents. Subsequently, the latents are then passed into the Inpainting UNet, which integrates features from the ID Extraction Module via self-attention operation together. Meanwhile, the reference body image is processed by CLIP image encoder to extract features, enabling semantic integration via cross-attention in both the ID Extraction Module and the Inpainting UNet. After denoising, the model outputs a resulting video that replaces the body in the target video with the reference body.}
    \label{fig:method_model_overview}
\end{figure*}
\label{method}

In this section, we begin with a brief introduction to preliminaries on Stable Diffusion in Sec.~\ref{method_diffusion}. Subsequently, Sec.~\ref{method_model} introduces each component of the SwapAnyone, detailing its role and the corresponding consistency. Finally, in Sec.~\ref{method_brighttrain}, we introduce the EnvHarmony strategy to achieve improved environment consistency.

\subsection{Preliminaries on Stable Diffusion}
\label{method_diffusion}
Stable Diffusion~\cite{stable_diffusion} transfers the diffusion process of DDPM~\cite{ddpm} from the pixel space to the latent space with a pair of Encoder $\mathcal{E}$ and Decoder $\mathcal{D}$. For an image $\boldsymbol{X}$, it employs the encoder $\mathcal{E}$ to first encode the image into a latent code $\boldsymbol{z}_0=\mathcal{E} \left(\boldsymbol{X} \right)$, then progressively adds Gaussian noise to the latent code through the forward diffusion process:
\begin{equation}
    q\left(\boldsymbol{z}_t|\boldsymbol{z}_{t-1}\right)=\mathcal{N}\left(\boldsymbol{z}_t;\sqrt{1-\beta_t}\boldsymbol{z}_{t-1},\beta_t\boldsymbol{\mathrm{I}}\right),
\end{equation}
where $t = 1, ..., T$, denotes the timesteps, $\beta_t \in \left(0, 1\right)$ is a
predefined noise schedule, and $\boldsymbol{\mathrm{I}}$ denotes identity matrix. Through a parameterization trick, $\boldsymbol{z}_t$ can be directly sampled from $\boldsymbol{z}_0$:
\begin{equation}
q\left(\boldsymbol{z}_t|\boldsymbol{z}_0\right)=\mathcal{N}\left(\boldsymbol{z}_t;\sqrt{\bar{\alpha}_t}\boldsymbol{z}_0,(1-\bar{\alpha}_t)\boldsymbol{\mathrm{I}}\right),
\end{equation}
where $\bar{\alpha_t}=\prod_{i=1}^t\alpha_i$ and $\alpha_t=1-\beta_t$. Diffusion models use a neural network $\epsilon_\theta$ to predict the added noise $\epsilon$ by minimizing the mean square error of the added noise and the predicted noise:
\begin{equation}
\min_\theta\mathbb{E}_{\boldsymbol{z},\epsilon\thicksim\mathcal{N}\left(0,\boldsymbol{\mathrm{I}}\right),t}\left[\|\epsilon-\epsilon_\theta\left(\boldsymbol{z}_t,t,\boldsymbol{c}\right)\|_2^2\right],
\end{equation}
where $\boldsymbol{c}$ is text prompt embedding or other conditions. During inference, a reverse Markov process is performed:
\begin{equation}
    p_\theta\left(\boldsymbol{z}_{t-1}|\boldsymbol{z}_t\right)=\mathcal{N}\left(\boldsymbol{z}_{t-1};\mu_\theta\left(\boldsymbol{z}_t,t\right),\Sigma_\theta\left(\boldsymbol{z}_t,t\right)\right),
\end{equation}
where $\mu_\theta\left(\boldsymbol{z}_t,t\right)$ and $\Sigma_\theta\left(\boldsymbol{z}_t,t\right)$ are parametrized by $\epsilon_\theta$. The reverse process begins with $\boldsymbol{z}_t$ sampled from $\mathcal{N} \left(0, 1\right)$. By decoding the final result $\hat{\boldsymbol{z}}_0$ of the reverse process, we can output image $\mathcal{\hat{\boldsymbol{X}}}$ conditioned on $\boldsymbol{c}$.
\subsection{End-to-End SwapAnyone Model}
\label{method_model}
Among the three consistencies in video body-swapping, identity consistency is primary goal. It evaluates whether the generated body's identity matches the reference, which requires SwapAnyone to extract the reference's identity features and integrate them into the generation process. Motion consistency requires the body's actions in the resulting video to align with those in the target video while maintaining smoothness, which necessitates SwapAnyone to understand motion information. Environment consistency requires the background of the resulting video to remain identical to that of the target video while ensuring visual harmony between the foreground body and the background, particularly in terms of luminance, which demands that SwapAnyone distinguishes luminance and identity to naturally integrate the reference’s identity with the luminance of the background.

The Inpainting UNet of SwapAnyone is built upon Stable-Diffusion-v1.5-Inpainting~\cite{stable_diffusion}. To support video input, the Inpainting UNet incorporates Temporal Layers. Additionally, SwapAnyone includes an ID Extraction Module that extracts identity information from the reference body, as well as a Motion Control Module responsible for encoding motion information. Moreover, SwapAnyone utilizes the image embedding of the reference body image produced by the CLIP image encoder instead of the text embedding generated by the CLIP text encoder, which is injected into the ID Extraction Module and the Inpainting UNet through the cross-attention operation. The overview of SwapAnyone is illustrated in Fig.~\ref{fig:method_model_overview}.

\textbf{Inpainting UNet.}
In the image inpainting task, the user provides an image $\boldsymbol{X}$ and a binary mask $\boldsymbol{M}$. After the emergence of Stable Diffusion, several inpainting techniques have been proposed. Currently, the most widely used technique modifies the first convolution layer of the UNet to accept an additional masked image and a mask input. Specifically, given an background image $\boldsymbol{X_\mathrm{b}}$, it is first encoded by the Image Encoder $\mathcal{E}$ into $\tilde{\boldsymbol{X}}_\mathrm{b}=\mathcal{E}\left(\boldsymbol{X}_\mathrm{b}\right)$, which is then concatenated with the corresponding downsampled mask $\tilde{\boldsymbol{M}}$ and a sampled noise tensor $\boldsymbol{z}_t$ from a Gaussian distribution $\mathcal{N}\left(0, 1\right)$:
\begin{equation}
    \tilde{\boldsymbol{z}}_t = \left[ \boldsymbol{z}_t, \tilde{\boldsymbol{X}}_\mathrm{b}, \tilde{\boldsymbol{M}}\right],
\end{equation}
where $\left[\cdot\right]$ represents the concatenation operation. Then the concatenated tensor is fed into the UNet to perform the reverse diffusion process. In the video body-swapping task, the input $\boldsymbol{V}$ is the background of the target video and the mask $\boldsymbol{M}$ is a binary image sequence representing the areas occupied by people. The inpainting process involves simultaneous denoising of all pixels, granting the Inpainting UNet an inherent ability to preserve environmental consistency. Furthermore, any control conditions like identity features and motion features directly affect the final output, making end-to-end optimization feasible.

\textbf{Temporal Layers.} The vanilla Stable-Diffusion-v1.5-Inpainting model is designed for the image inpainting task and cannot perform inpainting on videos. To enable SwapAnyone to handle video input and perform video inpainting rather than applying inpainting frame by frame, we incorporate Temporal Layers into the model. Inspired by the Motion Module in AnimateDiff~\cite{animatediff}, we add transformer layers that perform self-attention only in the temporal dimension after each block of the inpainting model, which ensures temporal consistency in the resulting videos.

\textbf{ID Extraction Module.}
Since the vanilla Stable-Diffusion-v1.5-Inpainting model relies solely on text input for guiding inpainting, it struggles to accurately preserve the reference body's identity details. Therefore, we introduce an ID Extraction Module that captures fine-grained features of the reference body. The module is built on ReferenceNet~\cite{animateanyone}, which is a copy of the Stable-Diffusion-v1.5 model, but unlike the implementation in AnimateAnyone, our ID Extraction Module takes both the reference body image and its corresponding DWpose image as inputs.

Given the reference body image $\boldsymbol{X}_\mathrm{r}$ and corresponding DWpose image $\boldsymbol{X}_\mathrm{p}$, we first encode them using the image encoder $\mathcal{E}$ to obtain $\tilde{\boldsymbol{X}}_\mathrm{r}=\mathcal{E}\left( \boldsymbol{X}_\mathrm{r}\right)$ and $\tilde{\boldsymbol{X}}_\mathrm{p}=\mathcal{E}\left( \boldsymbol{X}_\mathrm{p}\right)$. These encoded features are then concatenated and passed into the ID Extraction Module. Furthermore, the hidden states of $\tilde{\boldsymbol{X}}_\mathrm{r}$ and $\tilde{\boldsymbol{X}}_\mathrm{p}$ are extracted before passing through the spatial self-attention operation of the ID Extraction Module, which serves as conditions for the generation process. Let ID Extraction Module be denoted as $\mathcal{F}_\mathrm{r} \left(\cdot;\theta^\mathrm{r}\right)$ where $\theta^\mathrm{r}$ is the weights of ID Extraction Module and the fine-grained features extracted via the ID Extraction Module be represented as the condition $\boldsymbol{y}_\mathrm{r}$. For any denoising step $t$, the above process can be expressed as:
\begin{equation}
    \boldsymbol{y}_\mathrm{r}=\mathcal{F}_\mathrm{r}\left( \left[ \tilde{\boldsymbol{X}}_\mathrm{r},\tilde{\boldsymbol{X}}_\mathrm{p} \right] ,t;\theta^\mathrm{r}\right).
\end{equation}
Subsequently, we concatenate $\boldsymbol{y}_\mathrm{r}$ with the features of the Inpainting UNet at each layer to perform the self-attention operation jointly, which can inject fine-grained identity features of the reference into the Inpainting UNet. Mathematically, this process can be expressed as:
\begin{equation}
    \begin{gathered}
    \mathrm{Attention}\left( \boldsymbol{Q},\boldsymbol{K},\boldsymbol{V},\boldsymbol{y}_{\mathrm{a}} \right) =\mathrm{Softmax} \left( \frac{\boldsymbol{QK}'^{\mathrm{T}}}{\sqrt{d}} \right) \boldsymbol{V}', \\ \boldsymbol{Q}=\boldsymbol{W}^{\boldsymbol{Q}}\boldsymbol{z}_t,\boldsymbol{K}'=\boldsymbol{W}^{\boldsymbol{K}}\left[ \boldsymbol{z}_t,\boldsymbol{y}_{\mathrm{a}} \right] ,\boldsymbol{V}'=\boldsymbol{W}^{\boldsymbol{V}}\left[ \boldsymbol{z}_t,\boldsymbol{y}_{\mathrm{a}} \right].
    \end{gathered}
\end{equation}
The ID Extraction Module extracts fine-grained identity information, achieving better identity consistency compared to traditional text-conditioned or image-semantic-based identity injection methods.

\textbf{Motion Control Module.} In prior works, ControlNet~\cite{controlnet} is commonly used to encode pose sequences, and some lightweight modules~\cite{ipadapter,animateanyone} have also demonstrated the ability to control poses effectively. As a lightweight implementation, our Motion Control Module consists of only a few convolutional layers that encode DWpose images corresponding to the body of video frames and directly add the encoded pose features to the latents. After training, the Inpainting UNet learns to interpret the modified latent distribution and incorporate motion information from the poses, ensuring our results achieve good motion consistency.

\begin{figure*}[t]
    \centering
    \includegraphics[width=0.95\linewidth]{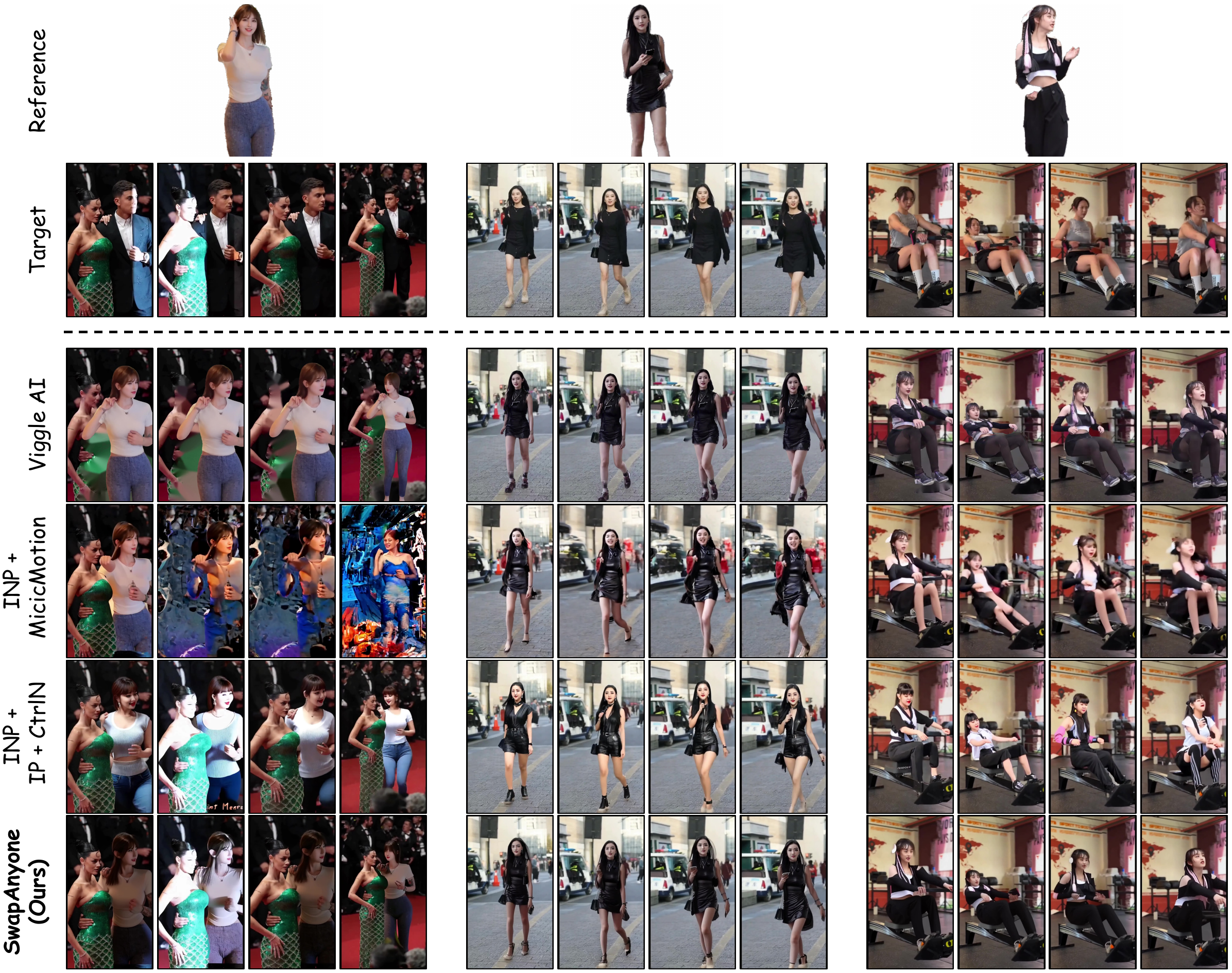}
    \caption{Comparison of visual quality across different methods. Viggle AI effectively preserves the identity of the reference body, but a noticeable boundary remains between the body and the background. Additionally, it struggles to handle occlusions between the body and objects in the background. The Inpainting model with MimicMotion struggles with background fidelity. The Inpainting model with IP Adapter and ControlNet lacks temporal modeling ability, leading to identity variations across frames. Our SwapAnyone maintains consistency of identity with the reference across frames while seamlessly blending the body with the background.}
    \label{fig:experiments_comparison}
\end{figure*}

\subsection{EnvHarmony Strategy}
Unlike controllable character video synthesis, video body-swapping allows identity differences between the reference body image and the target video. However, the final video must remain as visually harmonious as possible, regardless of these differences. 

Therefore, SwapAnyone requires that selectively utilizes the features provided by the reference, accepting those related to the identity, such as facial attributes and clothing details, while ignoring luminance. We adopt a straightforward strategy named EnvHarmony Strategy to achieve this.

The EnvHarmony strategy involves: Randomly adjust the brightness of the reference body image while keeping the brightness of the target video unchanged, or modify the brightness of the target video while preserving the brightness of the reference body image. During training, the reference and target originate from the same video, unlike in inference. Therefore, these brightness adjustments introduce luminance variations, helping to separate luminance features and emphasize identity features of the reference.
\label{method_brighttrain} 
\section{Experiments}
In this section, we will briefly introduce the construction of our dataset and present evaluation metrics for the video body-swapping task in Sec.~\ref{sec:dataset_and_eval}. Then, we present our experimental settings and compare SwapAnyone with other methods in Sec.~\ref{experiments_settings} and Sec.~\ref{experiments_comparison}, including evaluation metrics and user study results. Finally, in Sec.~\ref{experiments_ablation}, we provide an ablation study of key design choices in SwapAnyone.
\begin{table}[t]
    \centering
    \caption{Quantitative comparisons of different methods show that our approach achieves the best performance among all open-source and closed-source video body-swapping implementations.}
    \setlength{\tabcolsep}{2pt} 
    \begin{tabular}{@{}ccccc@{}}%
    \toprule
    & CLIP-Sim$\uparrow$ & OKS$\uparrow$ & MSE-B$\downarrow$ & FVD$\downarrow$ \\
    \midrule%
    Viggle AI &0.9523&0.73&18.31&165.17 \\
    INP + MimicMotion &0.9440&0.82&34.31&194.17 \\
    INP + IP + CtrN &0.9525&0.79&13.33&333.88 \\ 
    \textbf{SwapAnyone}&\textbf{0.9529}&\textbf{0.86}&\textbf{7.51}&\textbf{121.94} \\
    \bottomrule
    \end{tabular}
    \label{tab:experiments_comparison}
\end{table}
\subsection{Dataset and Evaluation}
\label{sec:dataset_and_eval}
We construct a video dataset named HumanAction-32K, sourced from the internet, which encompasses several categories, such as dance, sports, street footage, and daily vlogs, to cover a diverse range of human actions. Initially, we apply PySceneDetect~\cite{PySceneDetect} to segment the collected videos into clips, discarding those shorter than 3 seconds and retaining only vertical videos. Next, YOLO~\cite{yolo,yolov8_ultralytics} is employed to track people within the videos and identify the most frequently appearing ID. Finally, OpenPose~\cite{openpose} keypoints are extracted from frames using DWpose~\cite{dwpose} library, and videos are retained if frames with more than 11 visible key points constitute at least 75\% of the total frames.

Following common practices in video evaluation, we use Fréchet Video Distance (FVD)~\cite{fvd} as a metric for overall video quality. Moreover, the following metrics are introduced to effectively measure the quality of videos on the three types of consistency. These metrics evaluate on individual frames, while video results are calculated by averaging metrics across all frames.

For \textbf{identity consistency}, we select CLIP similarity as our metric, which is calculated using CLIP ViT-L/14~\cite{clip} to compare the body region of the reference body image with that of the resulting frame. For \textbf{motion consistency}, we use OKS (Object Keypoint Similarity), a metric from MS COCO~\cite{mscoco} designed to measure pose similarity. For \textbf{environment consistency}, we use the Mean Squared Error of Background (MSE-B) as a metric to evaluate background fidelity. Additionally, no existing metric can quantitatively evaluate the harmony of luminance in generated video. Therefore, we also use FVD as an effective metric for luminance consistency.

\subsection{Experiment Settings}
\begin{figure}
    \centering
    \includegraphics[width=0.9\linewidth]{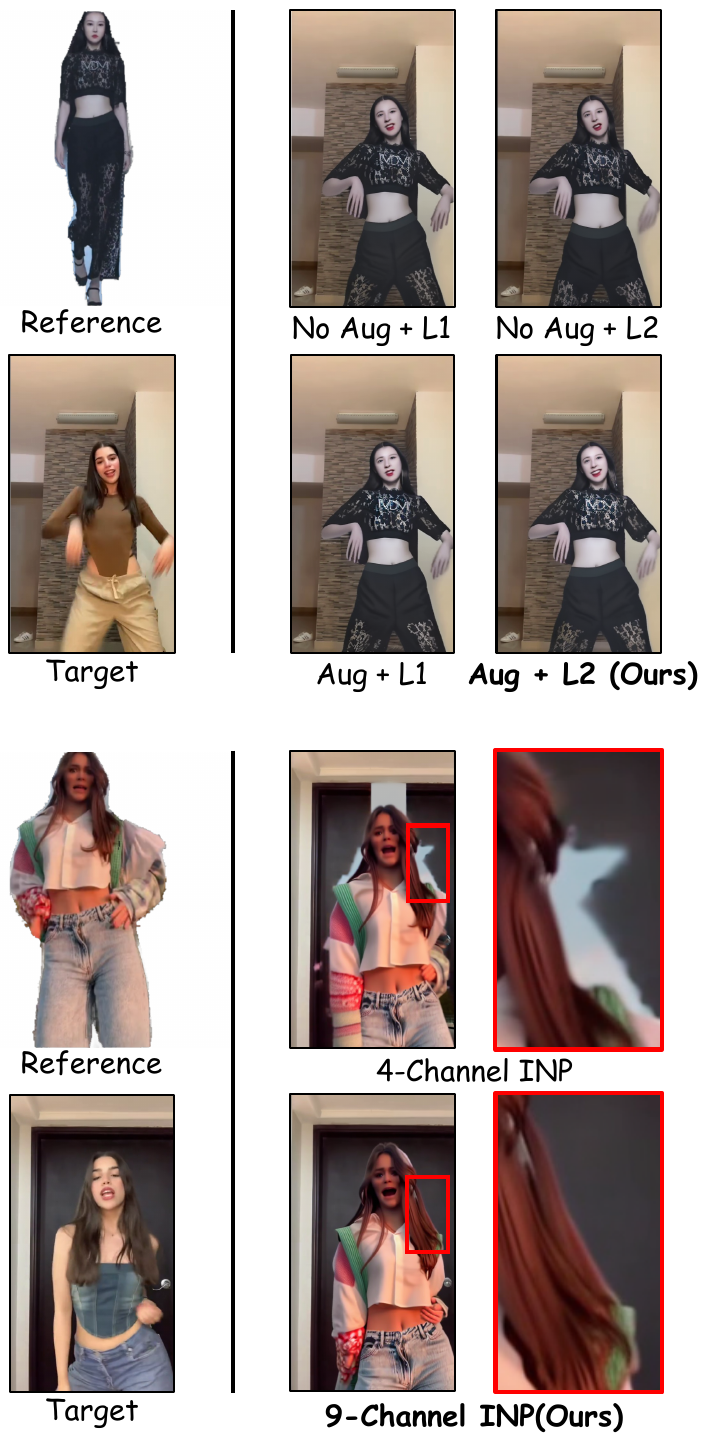}
    \caption{Ablation study visual comparison. The EnvHarmony strategy with data augmentation and MSE loss produces the most refined results. Additionally, the 9-channel Inpainting UNet achieves better background fidelity.}
    \label{fig:experiments_ablation}
\end{figure}
\label{experiments_settings}
We use Stable-Diffusion-v1.5-Inpainting as our Inpainting UNet. For other components, we initialize the ID Extraction Module using the ReferenceNet of AnimateAnyone~\cite{animateanyone}, the Motion Control Module using the Pose Guider of AnimateAnyone~\cite{animateanyone}, and the Temporal Layers using the Motion Module V2 of AnimateDiff~\cite{animatediff}. In addition, we replace the original CLIP text encoder with the image encoder of CLIP-ViT-L/14~\cite{clip}. During the training process, both the VAE and CLIP image encoder remain frozen. The training process of SwapAnyone consists of two stages: image pre-training stage and video fine-tuning stage.

At the image pre-training stage, the Temporal Layers are excluded. First, we extract two frames from a clip with an interval of at least 30 frames. One frame serves as the reference body image, where YOLO is used to extract body-related pixels while discarding others, and the other frame is used as the target. The processed reference body image and its corresponding DWpose image are fed into the ID Extraction Module, meanwhile, the DWpose image of the target is also input into the Motion Control Module. In this stage, we train for a total of 78k steps. During the first 30k steps, no additional adjustments are made to the data, allowing the model to gain basic inpainting ability using reference features. In the subsequent 48k steps, we incorporate the EnvHarmony strategy to enhance the ability of the model to decouple luminance and identity. Specifically, EnvHarmony adjusts brightness within a range of $\sigma$ times the original, where  $\sigma\in\mathcal{U}\left[0.5, 1.1\right]$, indicating that $\sigma$ follows a uniform distribution. During the entire process, we use a fixed resolution of $\text{768}\times\text{768}$ and batch size of 32, while the learning rate is configured at 1e-5.

At the video fine-tuning stage, we introduce the Temporal Layers and train on videos with the EnvHarmony strategy throughout the stage. To reduce the complexity of optimization, we freeze the ID Extraction Module, the Motion Control Module, and the spatial layers of the Inpainting UNet, optimizing only the newly added Temporal Layers. We select any frame from the video as the reference body image and the model predicts the entire video content with a total of 24 frames which are sampled at an interval of 4 frames. During the video fine-tuning stage, the learning rate is fixed at 1e-5, and we train for 30k steps at fixed resolution of $\text{768}\times\text{768}$ with batch size of 8.

\subsection{Comparison Results}
\label{experiments_comparison}
Although previous studies have not explicitly defined the video body-swapping task, the closed-source model Viggle AI~\cite{viggle} has demonstrated excellent video body-swapping performance. Among open-source models, several methods like Inpainting UNet with IP Adapter and ControlNet for image-based body-swapping already exist, which can be utilized to perform frame-by-frame image body-swapping to achieve video body-swapping. Additionally, the first frame can undergo image body-swapping, followed by pose-controlled image-to-video generation to achieve video body-swapping, with the Inpainting UNet with MimicMotion as a representative approach.

In our tests, Viggle AI utilizes the official v3-latest model, with IP Adapter and MimicMotion set to default. SwapAnyone employs DDIM~\cite{ddim} sampler with 30 steps, and Classifier-Free Guidance (CFG)~\cite{cfg} is set to 3.0. As shown in Tab.~\ref{tab:experiments_comparison}, SwapAnyone outperforms the other methods in three consistency metrics and FVD.
Fig.~\ref{fig:experiments_comparison} provides a visual comparison of these four methods. The closed-source model Viggle AI achieves strong identity consistency but shows a noticeable separation between the background and body while struggling with physical details like object occlusion. The Inpainting UNet with IP Adapter and ControlNet performs body-swapping frame by frame, leading to inconsistencies in identity across frames. Since the image-to-video approach lacks information about the background in subsequent frames of the target video, the Inpainting UNet with MimicMotion produces significant background deviations from the target video. In contrast, leveraging end-to-end training with the Inpainting UNet and the EnvHarmony Strategy, our model achieves seamless integration of the body and background while preserving realistic physical details.

Furthermore, we conduct a user study to evaluate the performance of the four methods. We select 10 high-resolution reference body images with fully visible bodies and high-resolution target videos containing one or more people. Participants are asked to select the best one among the four methods presented in random order based on four dimensions: identity consistency, motion consistency, environment consistency, and overall performance.

\begin{figure}
    \centering
    \includegraphics[width=\linewidth]{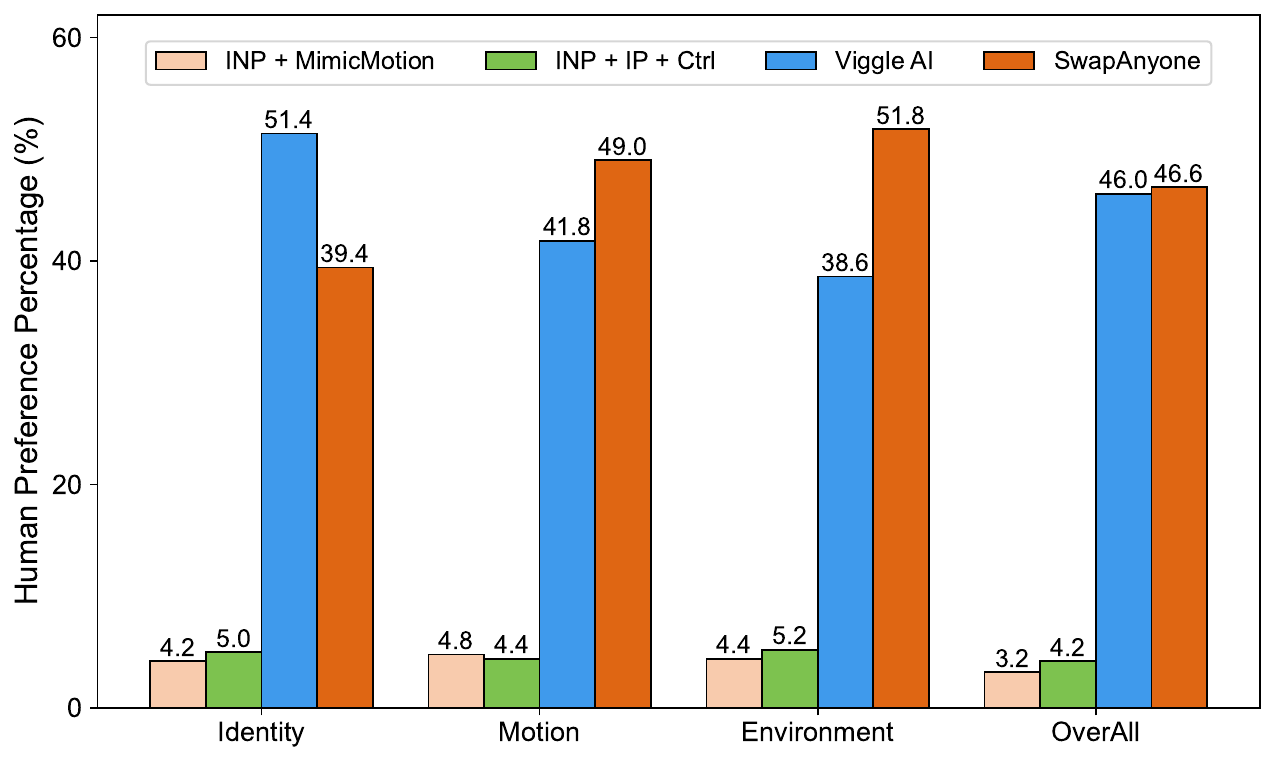}
    \caption{User study results of the four methods, where the Human Preference Percentage represents the proportion of best-performance evaluations each method receives in each dimension. SwapAnyone achieves performance comparable to the closed-source Viggle AI while surpassing other open-source models across all dimensions.}
    \label{fig:experiments_userstudy}
\end{figure}

Fig.~\ref{fig:experiments_userstudy} illustrates the user study results for various methods across the four evaluation dimensions. We collect a total of fifty responses from professionals and use the count of each method achieving the best performance as an evaluation metric. While SwapAnyone falls behind the closed-source model Viggle AI in identity consistency, it outperforms Viggle AI in motion consistency, environment consistency and gains comparable performance to Viggle AI in overall quality. Compared to open-source methods, SwapAnyone shows a distinct advantage in every dimension.

\subsection{Ablation Study}
\label{experiments_ablation}
\begin{table}[t]
    \centering
    \caption{Quantitative metrics of ablation study across different implementation methods in SwapAnyone. The 9-channel Inpainting UNet, combined with data augmentation and MSE loss, achieves the highest performance across all metrics. Therefore, we adopt this implementation for SwapAnyone.}
    \setlength{\tabcolsep}{5pt} 
    \begin{tabular}{@{}ccccc@{}}%
    \toprule
    & CLIP-Sim$\uparrow$ & OKS$\uparrow$ & MSE-B$\downarrow$ & FID$\downarrow$ \\
    \midrule%
    4-channel INP & 0.9423 & 0.77 & 15.01 & 44.83\\
    No Aug + MSE&0.9434&0.83&9.93&38.77 \\
    No Aug + MAE&0.9438&0.82&11.10&38.26 \\
    Aug + MAE&0.9436&0.83&9.86&38.42 \\
    \textbf{SwapAnyone}& \textbf{0.9441} & \textbf{0.84} & \textbf{9.85} & \textbf{37.83} \\
    \bottomrule
    \end{tabular}
    \label{tab:experiments_ablation}
\end{table}
We conduct an ablation study on several modules in SwapAnyone, including (1) the implementation of the EnvHarmony Strategy; and (2) the implementation of the Inpainting UNet. All test results are measured on 500 unseen samples, and the reported results are evaluated on the image pre-training stage.

\textbf{EnvHarmony Strategy.} We compare four implementations for the EnvHarmony strategy:
(1) Excluding data augmentation while using Mean Squared Error (MSE) loss; (2) Excluding data augmentation while using Mean Absolute Error (MAE) loss; (3) Using data augmentation and MAE loss; (4) Using data augmentation and MSE loss. Experiments are performed on image weights pre-trained for 30k steps without the EnvHarmony strategy, then further train for another 30k steps using different implementations. We report the differences among the four methods across various evaluation metrics, as shown in Tab.~\ref{tab:experiments_ablation}. The results indicate that the strategy of using data augmentation and MSE loss achieves the best performance across all dimensions. Fig.~\ref{experiments_ablation} presents a visual comparison of the different methods, illustrating that the strategy of using data augmentation and MSE loss effectively merges the identity features of the reference with the luminance of the target video.

\textbf{Inpainting UNet.} We compare two implementations: (1) inpainting without modifying the original UNet channels, where the background in the latents is replaced during each denoising step, and denoising is applied only to the masked region; (2) inpainting with five additional channels in the original UNet, where four channels are dedicated to background latents, and one channel is used for the mask. We first train for 30k steps without the EnvHarmony strategy using two inpainting implementations. Then, we enable the EnvHarmony strategy and continue training for another 30k steps. As demonstrated in Tab.~\ref{tab:experiments_ablation}, the approach employing nine concatenated channels outperforms the alternative, achieving better results across evaluation metrics. As shown in Fig.~\ref{fig:experiments_ablation}, the four-channel Inpainting approach struggles to inpaint regions around the human edges effectively, whereas the nine-channel method successfully integrates the body with the background.
\section{Conclusion}
In this work, we define the video body-swapping task and outline three corresponding consistencies: identity consistency, motion consistency, and environment consistency. Focusing on these consistencies, we propose a model structure called SwapAnyone, which includes Inpainting UNet, Temporal Layers, ID Extraction Module, and Motion Control Module. Additionally, we explore EnvHarmony Strategy, which significantly enhances environment consistency, especially luminance consistency. Finally, we introduce a dataset named HumanAction-32K for video body-swapping to facilitate further research in the community. Experiments demonstrate that our method achieves superior performance on the video body-swapping task.

{
    \small
    \bibliographystyle{ieeenat_fullname}
    \bibliography{main}
}

\end{document}